\documentclass[article]{IEEEtran}
\usepackage{cite}
\usepackage{amsmath,amssymb,amsfonts}
\usepackage{algorithmic}
\usepackage{graphicx}
\usepackage{booktabs}      
\usepackage{multirow}
\usepackage{subcaption}
\usepackage{textcomp}
\usepackage{url}
\usepackage{tikz}
\usepackage{hyperref}

\hypersetup{
    colorlinks=true,
    linkcolor=blue,
    filecolor=magenta,      
    urlcolor=blue,
    pdfpagemode=FullScreen,
    }

\def\BibTeX{{\rm B\kern-.05em{\sc i\kern-.025em b}\kern-.08em
    T\kern-.1667em\lower.7ex\hbox{E}\kern-.125emX}}

\begin{document}
\title{Improved skin lesion recognition by a Self-Supervised Curricular Deep Learning approach}
\author{Kirill Sirotkin, Marcos Escudero-Viñolo, Pablo Carballeira, Juan C. SanMiguel 
\thanks{This work has been submitted to the IEEE for possible publication. Copyright may be transferred without notice, after which this version may no longer be accessible. This work was supported by the Consejería de Educación e Investigación of the Comunidad de Madrid under Project SI1/PJI/2019-00414}
\thanks{Kirill Sirotkin, Marcos Escudero-Viñolo, Pablo Carballeira and Juan Carlos San Miguel are with the Escuela Politécnica Superior, Universidad Autónoma de Madrid, 28049, Madrid, Spain (e-mails: \{kirill.sirotkin, marcos.escudero, pablo.carballeira, juancarlos.sanmiguel\} @uam.es).}
}

\maketitle
\thispagestyle{empty}
\pagestyle{plain}


\bibliographystyle{IEEEtran}

\begin{abstract}

State-of-the-art deep learning approaches for skin lesion recognition often require pretraining on larger and more varied datasets, to overcome the generalization limitations derived from the reduced size of the skin lesion imaging datasets. ImageNet is often used as the pretraining dataset, but its transferring potential is hindered by the domain gap between the source dataset and the target dermatoscopic scenario. In this work, we introduce a novel pretraining approach that sequentially trains a series of Self-Supervised Learning pretext tasks and only requires the unlabeled skin lesion imaging data. We present a simple methodology to establish an ordering that defines a pretext task curriculum. For the multi-class skin lesion classification problem, and ISIC-2019 dataset, we provide experimental evidence showing that: i) a model pretrained by a curriculum of pretext tasks outperforms models pretrained by individual pretext tasks, and ii) a model pretrained by the optimal pretext task curriculum outperforms a model pretrained on ImageNet. We demonstrate that this performance gain is related to the fact that the curriculum of pretext tasks better focuses the attention of the final model on the skin lesion. Beyond performance improvement, this strategy allows for a large reduction in the training time with respect to ImageNet pretraining, which is especially advantageous for network architectures tailored for a specific problem.
\end{abstract}

\section{Introduction}
\label{sec:introduction}









According to the American Institute for Cancer Research, skin cancer is one of the most commonly occurring cancers in humans, and affects more than one million people worldwide every year \cite{bib_cancer_report}. Timely cancer diagnosis significantly increases the patient's chances of successful recovery. Therefore, it is of paramount importance to recognize the developing cancer in its early stages. However, many malignant skin lesions might go unnoticed if primary care clinicians do not recognize them as such, due to their lack of expertise or limited diagnosis time. Recent developments in AI-based automated assessment tools led to the creation of systems with significant potential for aiding medical analysis, as acknowledged by the latest EU legislative acts \cite{bib_coordinated_plan}. In particular, AI-based systems can improve skin lesion diagnostics by signaling if qualified medical personnel is necessary for diagnosis. In this regard, methods based on dermatoscopic image analysis, that do not require biopsying the lesion, are the preferred choice for early signaling.

 \begin{figure}[t]
  \centering
  \begin{subfigure}[t]{0.15\textwidth}
       \includegraphics[width=\textwidth]{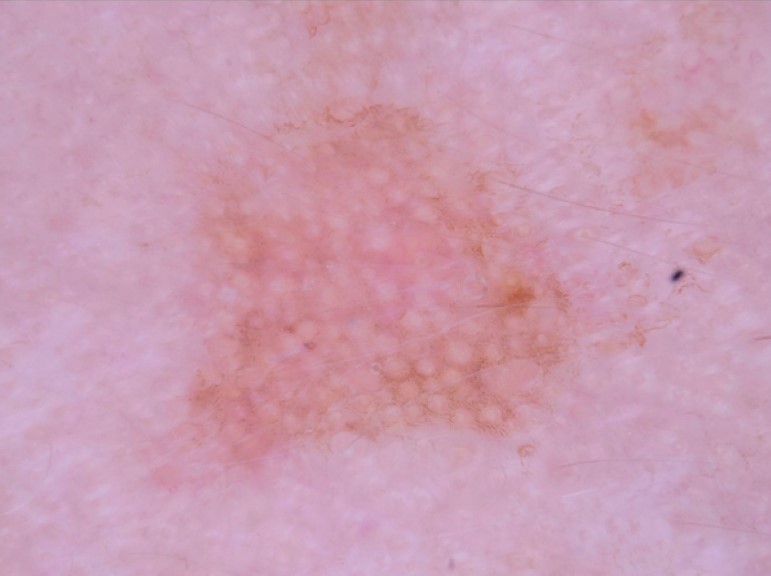}
  \end{subfigure}
  ~
  \begin{subfigure}[t]{0.15\textwidth}
       \includegraphics[width=\textwidth]{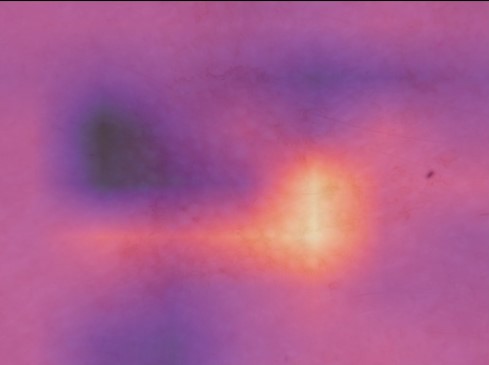}
  \end{subfigure} 
  ~
  \begin{subfigure}[t]{0.15\textwidth}
       \includegraphics[width=\textwidth]{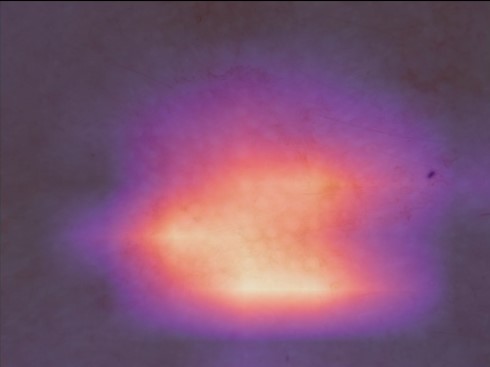}
  \end{subfigure} 
  \par\bigskip
  \begin{subfigure}[t]{0.15\textwidth}
       \includegraphics[width=\textwidth]{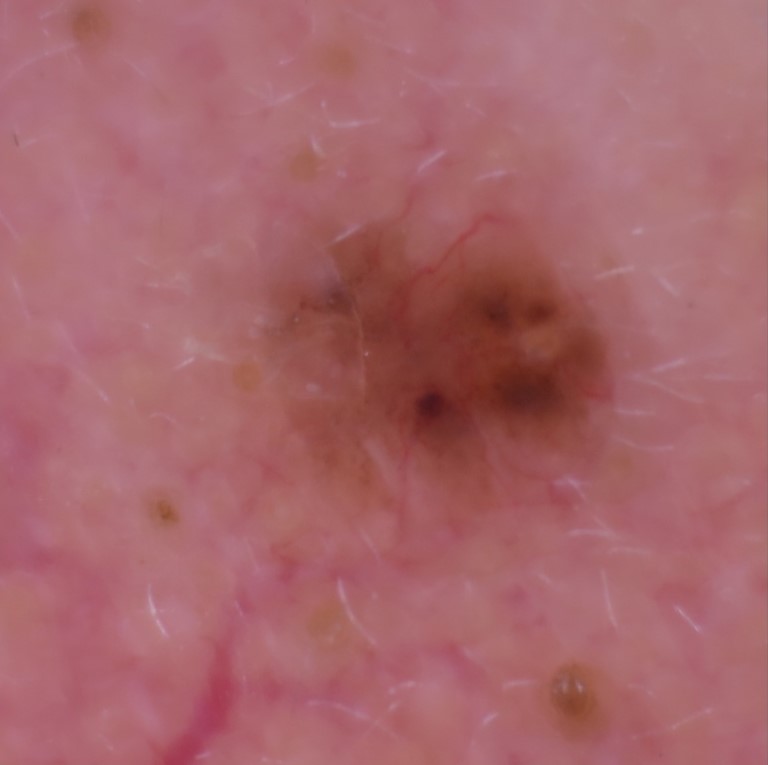}
       \caption{Original}
  \end{subfigure}
  ~
  \begin{subfigure}[t]{0.15\textwidth}
       \includegraphics[width=\textwidth]{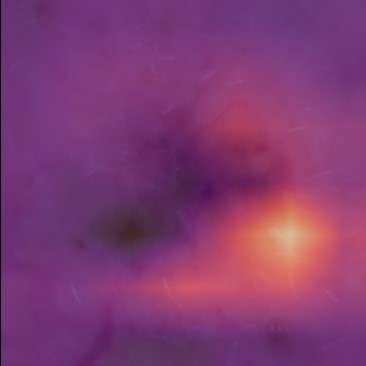}
    \caption{Anti-curriculum}
  \end{subfigure}
  ~
  \begin{subfigure}[t]{0.15\textwidth}
       \includegraphics[width=\textwidth]{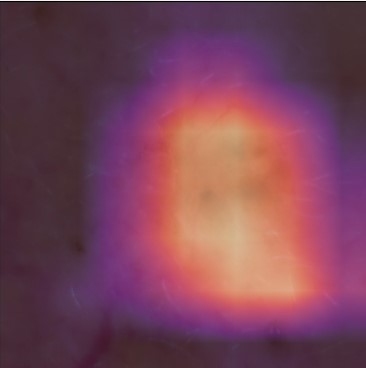}
       \caption{Curriculum}
  \end{subfigure}
  \caption{Class activation maps (the brighter the more relevant the corresponding region in the image is for the model prediction) of classifiers pre-trained with pretext tasks ordered in an advantageous (curriculum) and disadvantageous (anti-curriculum) fashion. Each row corresponds to a different sample image. 
}
  \label{fig_triplets}
\end{figure}
As of today, leading approaches for image-based skin lesion assessment \cite{bib_isic19_winner} are based on Convolutional Neural Networks (CNNs), achieving up to 72.5\% of accuracy on the multi-class classification problem of the popular benchmark dataset\textemdash ISIC-2019 \cite{bib_ham10000, isic_part_2, bib_bcn20000}, in which skin lesions are to be classified among 8 categories (i.e., melanoma, melanocytic nevus, vascular lesion, etc.). Traditionally, these solutions are trained in a supervised learning setting where the weights of the network are initialized randomly and automatically optimized to correctly predict the labels of the data in the training set by minimizing the value of the loss function that measures correctness of model's predictions \cite{bib_sgd}.
This strategy poses three restrictions on the training dataset\textemdash it must be labeled, large in size and diverse \cite{bib_impact_of_size}, so that the features learned from the training data can accurately approximate those found in the validation set. For its application to dermatoscopic imaging analysis\textemdash and generally medical imaging, these conditions entail a significant obstacle, as the availability of images themselves is limited due to the privacy restrictions and rareness of certain medical conditions. Moreover, even if the data is available, annotating it requires the skills and expertise of qualified doctors, which further contributes to the difficulty of dataset creation. For these reasons, some of the largest available medical imaging datasets consist of just around 100 000 labeled images (NIH Chest X-Ray dataset \cite{bib_nih}), while most others (PAD-UFES-20 \cite{bib_pad}, TCGA-LUAD \cite{bib_ct}, Lumbar Spine MRI \cite{bib_mri}, etc.) are significantly smaller, sometimes containing only a few hundred images per dataset\textemdash tens to hundreds times less than ImageNet \cite{bib_imagenet}. This makes supervised CNN training on such data largely ineffective.


To address the lack of data, it is beneficial to \textit{pretrain} CNN models on (preferably) domain-similar datasets to obtain a better starting point for training on the \textit{downstream task} (i.e., the target task, such as skin lesion classification) \cite{bib_how_transferable_feats}. Usually, such starting points are obtained by supervised training of a CNN model on a large and widely accepted as representative dataset, such as ImageNet~\cite{bib_imagenet}. Importantly, even when the domains of datasets used for pretraining and downstream task training differ significantly, pretraining still yields better performance than using randomly initialized weights~\cite{bib_how_transferable_feats}. Nonetheless, this limits the model selection to architectures with publicly available ImageNet-pretrained models, implying a computationally expensive and time consuming training procedure for new model architectures, or architectures designed ad-hoc for specific tasks, such as skin lesion recognition.




In such situations, a common method to pretrain a CNN model is to use Self-Supervised Learning (SSL)\textemdash a subset of unsupervised learning methods that leverages automatically generated labels as training objectives. The SSL-driving tasks, known as \textit{pretext tasks}, vary in their formulation and aim at solving problems that do not require manually annotated data: e.g., predicting randomly applied rotations \cite{bib_rotation}, colorization \cite{bib_colorization} or solving jigsaw puzzles \cite{bib_jigsaw}. Previous works show the advantages of SSL-pretraining applied for object recognition \cite{bib_moco_v1}, where SSL-pretrained models outperform models pretrained on ImageNet in a supervised regime, and skin lesion assessment \cite{bib_ssl_robustness}, where SSL-pretraining makes models more robust to noise.





In this study we take a step further for SSL pretraining by showing that the consecutive use of properly ordered pretext tasks significantly improves the results on ISIC-2019, with respect to pretraining using individual pretext tasks. Curriculum learning strategies \cite{bib_curriculum_learning}  propose to order samples during training according to their learning outcomes. Inspired by these techniques, we propose orderings of pretext tasks. Specifically, we hypothesize that, given a downstream task, an advantageous ordering of pretext tasks can be obtained by sorting the tasks according to the increasing order of their individual performances. We call orderings following this rule curriculum ones. Experimental results demonstrate that curriculum orderings tend to achieve higher performance on the downstream task than the rest of the possible orderings of the explored pretext tasks.



In addition to the analysis of their accuracies, we provide visual evidence that automatically obtained curriculum orderings of pretext tasks results in more effective internal representations that better focus on the skin lesion (see an example in Figure~\ref{fig_triplets}), which explain the increase in classification accuracy.



Overall, our analysis results in the following contributions:

\begin{itemize}
    \item We show that if pretext tasks are applied sequentially, their ordering has a significant effect on the model's accuracy after fine-tuning.
    \item Results indicate that curriculum orderings of the evaluated pretext tasks benefit the transferred performance over models not following these orderings.
    \item We demonstrate that\textemdash given a common architecture and using curriculum orderings, sequential SSL pretraining on ISIC-2019 dataset can outperform both supervised pretraining on ImageNet and the current winner of the ISIC-2019 challenge \cite{bib_isic19_winner}, reaching 75.44\% of balanced accuracy after fine-tuning for classification, while requiring significantly less time and resources for training.
\end{itemize}



\section{Related work} \label{Sec:related}

This section presents a brief overview of the current state-of-the-art in skin lesion recognition and discusses the use of Self-Supervised and Supervised Learning and in the presence of limited training data, specifically, medical imaging data.

\subsection{Limitation of traditional approaches to image recognition}

Most image recognition approaches are based on CNNs. A common strategy to improve a CNN's prediction accuracy on a given task is to increase the complexity of the model. This is done either by significantly changing the underlying network's architecture (i.e., AlexNet \cite{bib_alexnet} vs. VGG \cite{bib_vgg} vs. ResNet \cite{bib_resnet}) or adjusting its depth \cite{bib_resnet} and/or width \cite{bib_wide_net}. However, despite the growing performance on the ImageNet challenge, such approaches do not address the large limitation of any CNN, that is especially relevant to medical imaging datasets: their poor ability to generalize and learn efficient representations with limited training data.

Alternatively to changing the models themselves, one might resort to using ensembles of various networks or the same network trained multiple times (trials) with different initializations \cite{bib_ensembles_in_ml, bib_ensembles_nature}. However, this approach is very computationally expensive (as it requires to train the multiple models that comprise an ensemble, and later collect each model's prediction during inference). Furthermore, this paradigm is also not trivial to adapt to new tasks/domains, as it may imply the adaption of every model in the ensemble. 

\subsection{Current state of skin lesion recognition}

Due to privacy restrictions, to the relatively low rate at which certain medical conditions develop, and to the uneven distribution of suitable capture devices worldwide, a very small amount of medical imaging data is publicly available. In the subfield of skin lesion analysis, one of the largest accessible collections of imaging data is the International Skin Imaging Collaboration (ISIC) dataset that aggregates the data from other sources, such as HAM\_10000 dataset \cite{bib_ham10000}, MSK dataset \cite{isic_part_2} and BCN\_20000 dataset \cite{bib_bcn20000}. Annual public challenges based on the ISIC dataset target multi-class (predicting the exact type of a skin lesion) or binary (malignant vs. benign) classification problems. Whereas the latter can be considered an almost solved problem (ISIC-2020\textemdash a binary melanoma recognition problem\textemdash challenge winners achieved 0.949 on ROC AUC metric), the multi-class problem proposed in ISIC-2019 is still an open one, where the best reported approach \cite{bib_isic19_winner} currently reaches only 72.5\% $\pm$ 1.7\% of \textit{balanced accuracy} (average per-class accuracy).

Although ISIC-2019 is the largest publicly available skin lesion dataset with multi-class annotations, containing over 25000 labeled images, its size is small compared with those used for standard CNN training in well established tasks. Moreover, the number of samples per class ranges from 239 to 12875, making it highly imbalanced and further complicating CNN training\textemdash e.g., a vanilla ResNet-50 \cite{bib_resnet} does not reach 50\% of balanced accuracy, as shown in Table \ref{table_combinations_no_kd}. Some works address this issue by designing new loss functions that account for severe class imbalances \cite{bib_isic_loss_functions}. However, the general trend seems to be increasing the complexity of neural models and utilizing deeper architectures, such as DenseNets \cite{bib_densenet} or very deep ResNets \cite{bib_resnet}. Continuing in the same direction, the top three best performing approaches in ISIC-2019 skin lesion diagnosis challenge are based on ensembles of neural networks that leverage multiple models to infer predictions  \cite{bib_isic19_winner, bib_isic_19_winner_2, bib_isic_19_winner_3}. 


\subsection{Transfer learning and Self-Supervised Learning}


A different and popular approach to increase the classification accuracy of deep learning architectures is transfer learning: using a model trained for one task as a starting point for training for a different task. Recent studies introduce strategies for pretraining models without relying on labelled data \cite{bib_rotation, bib_moco_v1}. This is achieved by using SSL-based methods that solve a \textit{pretext task} by formulating objectives that do not require manually labelled data. It is assumed that solving pretext tasks requires a high-level understanding of the data and, therefore, allows to learn efficient visual representations. Hereby, the features learned through solving \textit{pretext tasks} are considered to be starting points suitable for transfer learning. For instance, a well-known \textit{pretext task}\textemdash rotation prediction~\cite{bib_rotation}\textemdash generates training labels that encode a series of rotations applied to the input image, anticipating that the information required to correctly rotate an image requires the \textit{learning} of some fundamental object qualities.

\subsection{Curriculum learning}


Although the number of existing SSL strategies is constantly increasing, a categorization according to the underlying label generation strategy can be used to assign them to one of three categories: \textbf{geometric} (Rotation prediction \cite{bib_rotation}, Relative patch location \cite{bib_relative_location}, Jigsaw puzzles \cite{bib_jigsaw}),  \textbf{clustering} (Deep Clustering \cite{bib_dc}, Online Deep Clustering \cite{bib_odc}, ClusterFit \cite{bib_clusterfit}) and  \textbf{contrastive} (Momentum Contrast \cite{bib_moco_v1, bib_moco_v2}, Bootstrap Your Own Latent \cite{bib_byol}, Non-Parametric Image Discrimination \cite{bib_npid}, Simple Framework for Contrastive Learning of Visual Representations \cite{bib_simclr}). 

Relative patch location prediction, being a typical \textbf{geometric} SSL model, splits an input image into patches, then samples two adjacent ones and trains to predict their relative location. On the other hand, \textbf{clustering} SSL models, such as Online Deep Clustering (ODC), use classical clustering algorithms to generate intermediate training labels. Finally, \textbf{contrastive} models rely on the use of the contrastive loss function to discriminate between positive and negative samples. Often the negative samples are generated through the data augmentations of positive samples, as it is done in the case of Momentum Contrast (MoCo) model.

Currently, regarding the accuracies achieved using ImageNet as the downstream classification task, \textbf{contrastive} models outperform \textbf{geometric} and \textbf{clustering} ones: MoCo v3 \cite{bib_moco_v3}\textemdash81.0\%, SimCLR v2\cite{bib_simclr_v2}\textemdash79.8\% and BYOL\cite{bib_byol}\textemdash79.6\%, while the closest \textbf{clustering} model\textemdash DeepCluster v2\cite{bib_dc}\textemdash yields only 75.2\% of accuracy and a classic \textbf{geometric} model\textemdash Rotation prediction\textemdash reaches 55.4\% of accuracy.

\subsection{SSL in skin lesion recognition}

The promise of learning useful representations without requiring labelled data led to the applications of SSL strategies to the medical imaging, where data labelling is challenging. In particular, a number of recent works target the skin lesion recognition problem leveraging SSL approaches for model pretraining. This narrows the gap between SSL and fully supervised pretraining on ImageNet but, in most cases, does not yet close it. For example, a recently proposed approach \cite{bib_topology} reaches 80.6\% of accuracy on a multi-class ISIC-2018 classification problem by employing self-supervision to obtain transformation-invariant features. Specifically, features extracted at each epoch from the image decoder module of a CycleGAN architecture \cite{bib_cyclegan} are assigned to $N$ clusters without a prior knowledge of $N$, using the maximum modularity clustering algorithm \cite{bib_modularity_clustering}. The memberships of the samples are used as pseudo-labels to optimize the features. A different study compares the individual performances of five existing SSL models \cite{bib_ssl_evaluation_2021} (BYOL \cite{bib_byol}, MoCo \cite{bib_moco_v2}, SimCLR \cite{bib_simclr}, InfoMin \cite{bib_infomin}, SwAV \cite{bib_swav}) on the ISIC-2019 data, but only considers the binary classification problem, reaching 0.956 on the standard for binary problems metric\textemdash ROC AUC (area under the precision-recall curve). On the contrary, another recent work does consider the multi-class skin lesion classification problem and shows that SimCLR pretraining outperforms supervised pretraining \cite{bib_super_dataset}. However, this study relies on a private dermatoscopic dataset containing over 450 000 samples. Nevertheless, there is a gap in the current state-of-the-art as, to our knowledge, none of the existing approaches considers more than one SSL task in the pretraining stage, thus, limiting the resulting performance by not taking advantage of the pretext tasks of different nature. Moreover, with seldom exceptions, most works that use SSL pretraining in the skin lesion domain rely only on \textbf{contrastive} models (as these are generally the most accurate ones), thus, the potential contributions of \textbf{clustering} and \textbf{geometric} models are still barely explored. Finally, a multi-class skin lesion recognition problem is relatively unexplored by SSL methods, as the majority of works focus on the binary melanoma recognition task, and only a few studies target the multi-class problem of the older (and smaller) versions of the ISIC dataset. 



A promising method for improving the accuracy of deep learning architectures emerges from the careful selection and ordering of training samples\textemdash curriculum learning \cite{bib_cl_starting_small, bib_shaping, bib_curriculum_learning}. The idea behind it is that, inspired by human learning processes, the training of deep learning architectures may benefit from ordering of data samples based on some evidence of their \textit{complexity}, e.g., their contribution in the training loss. 

Generally, a curriculum of samples starts by using for training those samples that are learned faster and progressively incorporates the rest. For visual tasks, reported results indicate that curriculum training results in models that yield better performance than those obtained when samples are selected at random \cite{bib_curriculum_learning}. However, despite the benefits of curriculum learning at the samples level, to our knowledge no study investigates the effects of a different orderings of models pre-trained with multiple sequential tasks. In this work we define a curriculum order of pretext tasks that optimizes the performance of a model pre-trained sequentially on those tasks. Finally, we investigate applications of curriculum ordering of pretext tasks and their effects on the skin lesion recognition performance.



\section{Proposed approach}
\label{sec:Methodology}
\begin{figure}[t]
    \centering
    \includegraphics[width=0.5\textwidth]{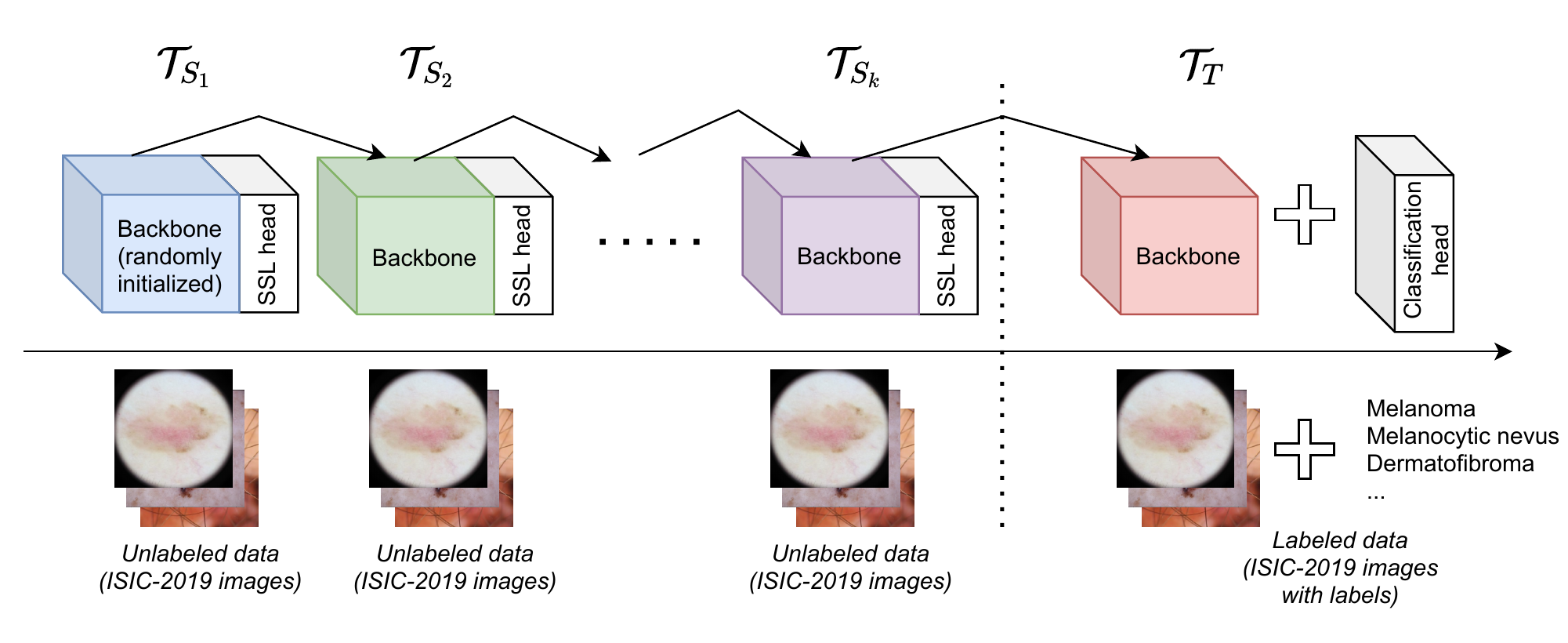}
    \caption{Proposed pretraining scheme.}
    \label{fig_ssl_ordering}
\end{figure}

\begin{table}
\centering
\caption{Accuracies for the evaluated single- and multi-source transfer settings for the ISIC-19 skin lession recognition task. The right-most column indicates whether the pretraining strategy led to a higher classification accuracy than supervised pretraining on ImageNet. The column "$\delta$" indicates how the performance of a combination of pretext tasks differs from an individual pretext task. The left-most column shows whether a combination follows Curriculum (C), Anti-Curriculum (AC) or Mixed Curriculum (MC) ordering.
}
\setlength{\tabcolsep}{2pt}
\label{table_combinations_no_kd}
\begin{tabular}{@{}rcccccc@{}}
\toprule
\multicolumn{1}{c}{} &
  1st task &
  2nd task &
  3rd task &
  \begin{tabular}[c]{@{}c@{}}Balanced \\ accuracy\\ (\%)\end{tabular} &
  $\delta$ (\%) &
  \begin{tabular}[c]{@{}c@{}}Better\\ than\\ ImageNet\end{tabular} \\ \midrule
-                    & Rel. loc.   &             &             & 69.52 & -     & No           \\
(AC)                 & Rel. loc.   & ODC         &             & 70.68 & 1.16  & No           \\
(MC)                 & Rel. loc.   & ODC         & MoCo v2     & 75.00 & 5.49  & \textbf{Yes} \\
(C)                  & Rel. loc.   & MoCo v2     &             & 74.10 & 4.58  & \textbf{Yes} \\
(MC)                 & Rel. loc.   & MoCo v2     & ODC         & 74.38 & 4.86  & \textbf{Yes} \\ \midrule
-                    & MoCo v2     &             &             & 72.74 & -     & No           \\
(AC)                 & MoCo v2     & ODC         &             & 72.72 & -0.02 & No           \\
(MC)                 & MoCo v2     & ODC         & Rel. loc.   & 67.00 & -5.74 & No           \\
(AC)                 & MoCo v2     & Rel. loc.   &             & 66.72 & -6.02 & No           \\
(AC)                 & MoCo v2     & Rel. loc.   & ODC         & 69.80 & -2.95 & No           \\ \midrule
-                    & ODC         &             &             & 63.52 & -     & No           \\
(C)                  & ODC         & Rel. loc.   &             & 68.23 & 4.71  & No           \\
(C)                  & ODC         & Rel. loc.   & MoCo v2     & 73.36 & 9.84  & No           \\
(C)                  & ODC         & MoCo v2     &             & 75.44 & 11.92 & \textbf{Yes} \\
(MC)                 & ODC         & MoCo v2     & Rel. loc.   & 65.73 & 2.21  & No           \\ \midrule
\multicolumn{1}{c}{} &
  \multicolumn{3}{c}{ISIC-2019 challenge winner \cite{bib_isic19_winner}} &
  72.5 $\pm$ 1.7 &
  - &
  - \\
\multicolumn{1}{c}{} & \multicolumn{3}{c}{Supervised ImageNet} & 73.76 & -     & -            \\
\multicolumn{1}{c}{} & \multicolumn{3}{c}{No pretraining}      & 49.27 & -     & -            \\ \bottomrule
\multicolumn{7}{p{251pt}}{}\\
\end{tabular}
\end{table}






In this study we propose to increase the accuracy on the ISIC-2019 multi-class problem by improving the pretraining pipeline, frequently applied in transfer learning approaches to the skin lesion recognition problem. The basic idea of our method (see Figure \ref{fig_ssl_ordering}) is to extend the recent works on SSL pretraining for skin lesion recognition and
leverage multiple, instead of a single, pretext tasks in the pretraining stage. 

\subsection{Preliminaries}

To clarify the proposed approach we first formalize the notation used in this work, following the nomenclature proposed in previous works \cite{bib_survey_tl}:


\vspace{0.2cm}

\textit{Domain}: A domain $\mathcal{D} = \{\mathcal{X}, P(X)\}$ is comprised of the complete feature space $\mathcal{X}$ (i.e., all images) and a marginal distribution $P(X)$, where $X=\{x|x_i \in \mathcal{X}, i=1,..., n\}$ is the set of images defining the domain (i.e., skin lesion images).

\vspace{0.2cm}

\textit{Task}: A task $\mathcal{T}=\{\mathcal{Y}, \mathcal{F}\}$ is comprised of the label space $\mathcal{Y}$ (i.e., melanoma, melanocytic nevus,...) and the implicit decision function $\mathcal{F}$ which is expected to be learned by the model from the input data (i.e., conditional probability of labels, given the samples).

\vspace{0.2cm}

\textit{Transfer learning}: Given (a) data distribution(s)  $\{\mathcal{D}_{S}, \mathcal{T}_{S_i}|i=1,..., m\}$ drawn from the \textit{source} ($S$) domain(s) and task(s), and a data distribution $\{\mathcal{D}_{T}, \mathcal{T}_{T}\}$ drawn from the \textit{target} ($T$) domain and task, transfer learning uses the knowledge from the source domain(s) to improve the performance of the decision function $\mathcal{F_T}$ on the target domain.

\vspace{0.2cm}

\textit{Single-source transfer learning}: Transfer learning setup in which $m=1$. In other words, the knowledge is transferred from a single source domain to the target domain. Most transfer learning approaches follow single-source scenario.

\vspace{0.2cm}

\textit{Multi-source transfer learning}: Transfer learning setup in which $m \ge 2$. In other words, the knowledge is transferred from multiple source domains to the target domain.

\vspace{0.2cm}

\subsection{Self-supervised Curricula}
With these definitions, we formulate the proposed approach as follows: given a single source domain and $k$ source tasks $\{\mathcal{D}, \mathcal{T}_{S_i}|i=1,..., k\}$, we define a multi-source transfer learning approach that uses the knowledge from $\{\mathcal{D}, \mathcal{T}_{S_i}\}$ to improve the performance of the learned target decision function $\mathcal{F}_T$. In this setting, the knowledge is sequentially transferred from $\{\mathcal{D}, \mathcal{T}_{S_{i}}\}$ to $\{\mathcal{D}, \mathcal{T}_{S_{i+1}}\}$ (the first task $\mathcal{T}_1$ is trained from scratch). At step $k$ we transfer the knowledge from $\{\mathcal{D}, \mathcal{T}_{S_k}\}$ to the target task $\{\mathcal{D}, \mathcal{T}_T\}$.







Note, that in our case the source $\mathcal{D}_S$ and the target $\mathcal{D}_T$ domains are the same, therefore, the whole process can be viewed as the sequential training of $k+1$ models on the source domain.

Currently, optimal ordering of $k$ pretext tasks remains an open question. We define the best performing ordering as the curriculum one and make a hypothesis that it coincides with the ordering $\mathcal{A}_{1}, \mathcal{A}_{2},... \mathcal{A}_{k}$ of accuracies obtained with each pretext task individually, where $\mathcal{A}_{1} \leq \mathcal{A}_{2} \leq ... \leq \mathcal{A}_{k}$. Further, in Section \ref{sec:Experiments}, we present experimental findings that support this hypothesis.



\section{Experiments}
\label{sec:Experiments}

\subsection{Setup}

\setlength{\tabcolsep}{4pt}

\begin{table}[t]
\centering
\caption{Comparison of SSL pretraining approaches on the ISIC-2017 binary classification challenge. Pretext-Invariant Representation Learning (PIRL) \cite{bib_pirl} with image transformations produced by the Jigsaw \cite{bib_jigsaw} and Rotation prediction \cite{bib_rotation} pretext tasks, along with individual Relative Location \cite{bib_relative_location}, ODC\cite{bib_odc} and MoCo v2\cite{bib_moco_v2} pretext tasks are outperformed by a combination of Relative Location, ODC and MoCo v2. ROC AUC stands for the area under the receiver operating characteristic curve, PR AUC stands for the area under the precision-recall curve.}
\label{table_isic_2017}
\begin{tabular}{@{}cccc@{}}
\toprule
Starting point & ROC AUC & PR AUC & Accuracy \\ \midrule
Random initialization & 0.598   & 0.257  & 79.67    \\
ImageNet              & 0.791   & 0.525  & 83.03    \\ \midrule
PIRL (Jigsaw) \cite{bib_ssl_isic_2017}         & 0.664   & 0.32   & 79.8     \\
PIRL (Rotation) \cite{bib_ssl_isic_2017}       & 0.732   & 0.37   & 80.3     \\ \midrule
Relative Location     & 0.663   & 0.33   & 78.5    \\
ODC                   & 0.696   & 0.37   & 80.33    \\
MoCo v2               & 0.731   & 0.4    & 80.16    \\ \midrule
\textbf{RL $\,\to\,$ ODC $\,\to\,$ MoCo v2 (ours)} & \textbf{0.754} & \textbf{0.48} & \textbf{82.3} \\ \bottomrule
\end{tabular}
\end{table}

We follow the method described in Section \ref{sec:Methodology} to assess the effects of sequential pretraining of multiple pretext tasks on the ISIC-2019 multi-class classification challenge (as the latest version of ISIC that addresses the multi-class problem). As consecutive learning of pretext tasks changes the learned representations, it can be expected that same-category pretext tasks learned consecutively will not change the learned representations significantly as they are optimized on the basis of similar objectives (i.e., the representations learned by the pretext tasks of the same category\textemdash \textbf{geometric}, \textbf{clustering} or \textbf{contrastive}\textemdash could be expected to be similar). Therefore, we pick three representative examples of each group of pretext tasks (Relative Location, MoCo v2, ODC) to ensure that the representations learned by them differ between each other, define curriculum orderings and compare them against all other possible combinations of them in the training pipeline. The following steps are used to define the order of pretext tasks:


\begin{enumerate}
    \item Given three pretext tasks, we perform single-source transfer learning with the same target and source domains and record the achieved accuracies: $\mathcal{A}_{RL}, \mathcal{A}_{ODC}, \mathcal{A}_{MoCo\_v2}$. All three pretext tasks use the same backbone architecture: ResNet-50\cite{bib_resnet}.
    

    \item Following the hypothesis outlined in Section \ref{sec:Methodology} we define as curriculum orderings those coinciding with the increasing single-source transfer learning accuracies (accuracy hereafter). We group all orderings as follows:
    \begin{itemize}
        \item Curriculum: ordered by increasing accuracy.
        \item Anti-curriculum: ordered by decreasing accuracy.
        \item Full (anti-)curriculum: (anti-)curriculum comprised by all tasks.
    
    \end{itemize}
 Any other ordering is considered a Mixed-curriculum.

\end{enumerate}



Given the individual accuracies of pretext tasks, we define all the curriculum orderings of two and three pretext tasks for the multi-source transfer learning and train corresponding models. Further, in Section \ref{results_main}, we show that these curriculum orderings are the best performing ones (see Tables \ref{table_combinations_no_kd} and \ref{table_isic_2017}), thereby, proving our hypothesis.

Training regimes and hyperparameters for the pretext tasks are presented in Table \ref{table_ssl_params}
. When transferring the knowledge to the target task, we select the checkpoint from the last pretext task in the pretraining pipeline at the epoch at which the best accuracy is achieved on the validation set (Relative Location) or we simply load the weights from the last epoch when validation set is not available due to the nature of the pretext task (ODC, MoCo v2).


As baselines for comparison we use a model trained from scratch on the target\textemdash skin-lesion recognition\textemdash task (No pretraining) and a model fine-tuned end-to-end on the target task but pre-trained on the ImageNet dataset (Supervised ImageNet). Training hyperparameters for these two models are compiled in the last two rows of Table \ref{table_ssl_params}




\subsection{Results for skin-lesion classification} \label{results_main}

\begin{figure}[t]
  \centering
       \includegraphics[width=0.5\textwidth]{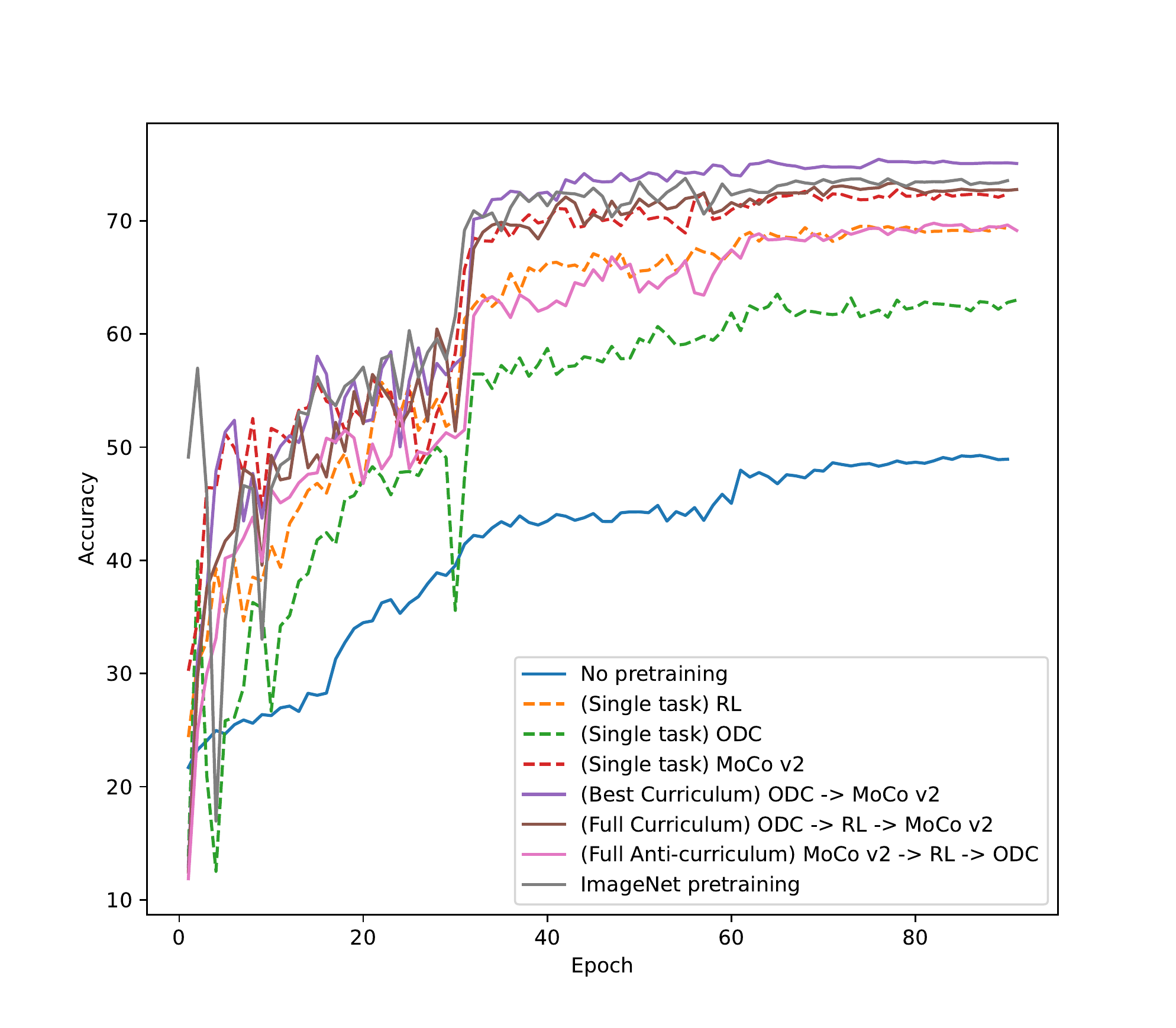}
       \caption{Validation accuracies of classifiers (ResNet-50 models) trained on the features learned with different pretext tasks (or their combinations).}
  \label{fig_accuracies}
\end{figure}


Table \ref{table_combinations_no_kd} presents the classification accuracies on the ISIC-2019 dataset achieved with different single-source and multi-source transfer learning setups. Given the classification accuracies obtained with the three single-source setups (first row of each block in Table~\ref{table_combinations_no_kd}): $\mathcal{A}_{ODC}=63.5$, $\mathcal{A}_{RL}=69.5$,  $\mathcal{A}_{MoCo\_v2}=72.7$, we define the full curriculum in the multi-source approach as ODC $\,\to\,$ Relative Location  $\,\to\,$ MoCo v2 and the full anti-curriculum as MoCo v2 $\,\to\,$ Relative Location $\,\to\,$ ODC. 




The results shown in Table \ref{table_combinations_no_kd} also demonstrate the effect of multi-source pretext task pretraining on the  classification accuracies on ISIC-2019. Specifically, when the pretraining begins with Relative Location or ODC, thereby avoiding the supposed anti-curriculum ordering, the resulting accuracy is always superior to single-source pretext task pretraining. 
On the contrary, orderings that are expected to be anti-curriculum (or mixed curriculum) starting with MoCo v2 (e.g., MoCo~v2 $\,\to\,$ ODC; MoCo~v2 $\,\to\,$ ODC $\,\to\,$ RL; MoCo~v2 $\,\to\,$ RL; MoCo~v2 $\,\to\,$ RL $\,\to\,$ ODC) always harm the performance of the individual model (MoCo~v2).





Moreover, five pretraining setups lead to state-of-the-art results after transfer learning, including one where only a single SSL model was trained (MoCo v2); two pretraining setups outperform the current ISIC-2019 winner\cite{bib_isic19_winner} (Relative Location $\,\to\,$ ODC $\,\to\,$ MoCo v2 and ODC $\,\to\,$ MoCo v2); and four pretraining setups outperform the results achieved with fully supervised ImageNet pretraining, with the best result of 75.44\% achieved with a pretraining pipeline of two pretext tasks following a curriculum ordering (ODC $\,\to\,$ MoCo~v2). Overall, combinations of pretext tasks following a curriculum ordering outperform the individual models in all cases and reach the highest performances. Hence, the results support our definition of curriculum.

Figure \ref{fig_accuracies} also demonstrates the accuracy curves on the target task for models pretrained with different setups and reveals that more accurate models do not only reach higher performances but also do it faster than less accurate ones.

To our knowledge, no work targeting a multi-class problem on ISIC-2019 dataset used SSL methods. Therefore, we additionally evaluate our approach on an older version of the ISIC dataset\textemdash ISIC-2017 \cite{bib_isic2017_dataset}\textemdash where SSL approaches were tested before. Table \ref{table_isic_2017} summarizes the results and demonstrates that while none of the three SSL models tested by us (Relative Location, ODC and MoCo v2) outperforms previously published results, their curriculum combination does.

\begin{figure}[t]
  \centering
       \includegraphics[width=0.49\textwidth]{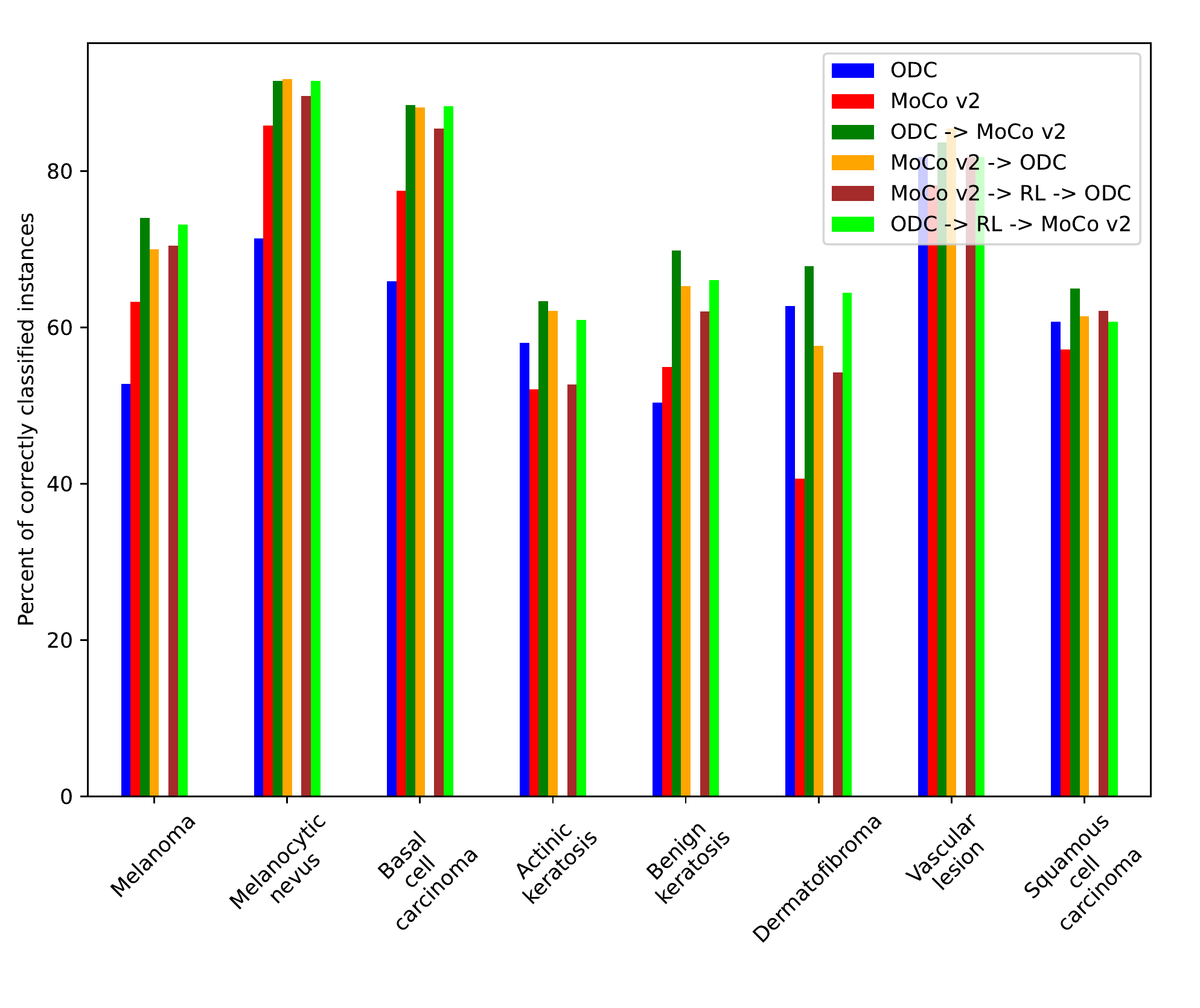}
  
  \caption{Percentage of correctly classified images per class by models with different pretraining strategies.}
  
  \label{fig_bars}
\end{figure}

Finally, the bar chart shown on Figure~\ref{fig_bars} indicates that the accuracy gains achieved by different pretraining setups are evenly distributed among all categories of ISIC-2019 and do not come from a large increase in accuracy in only one or few categories:  combinations following the curriculum ordering (ODC $\,\to\,$ MoCo~v2 and ODC $\,\to\,$ RL $\,\to\,$ MoCo~v2) consistently improve the results of the two individual pretext tasks (ODC and MoCo~v2).


\subsection{Qualitative analysis of the results}

\begin{figure*}[t]
  \centering
  \begin{subfigure}[t]{0.17\textwidth}
       \includegraphics[width=\textwidth]{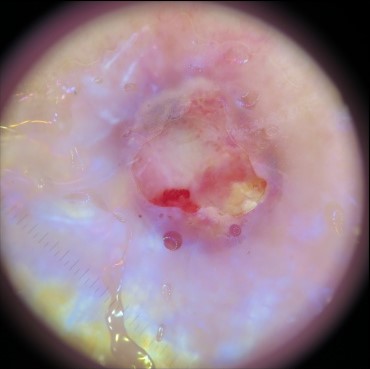}
  \end{subfigure}
  ~
  \begin{subfigure}[t]{0.17\textwidth}
       \includegraphics[width=\textwidth]{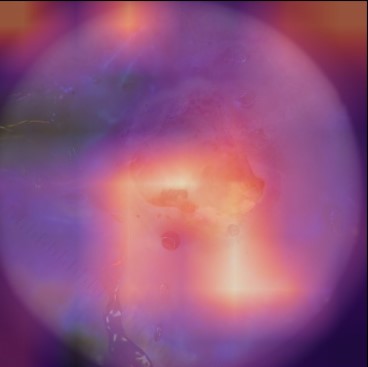}
       \label{fig_no_ssl_no_ssl}
  \end{subfigure} 
  ~
  \begin{subfigure}[t]{0.17\textwidth}
       \includegraphics[width=\textwidth]{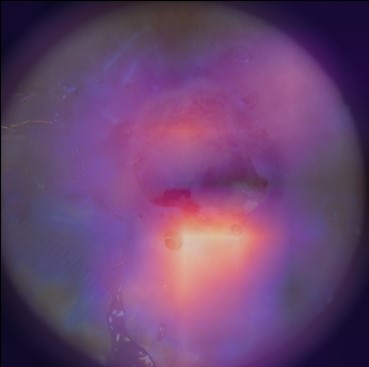}
  \end{subfigure}
  ~
  \begin{subfigure}[t]{0.17\textwidth}
       \includegraphics[width=\textwidth]{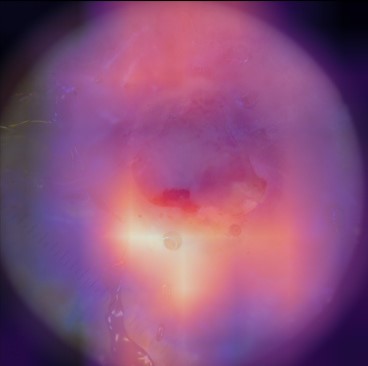}
  \end{subfigure}
  ~
  \begin{subfigure}[t]{0.17\textwidth}
       \includegraphics[width=\textwidth]{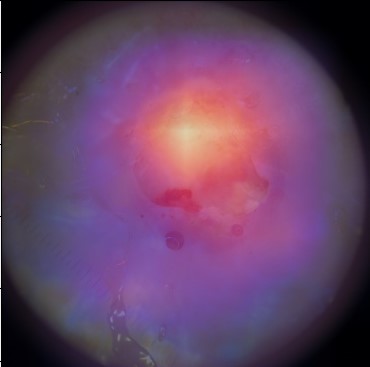}
  \end{subfigure} 
  \hfill
  \\
  \vspace{-12pt}
  \begin{subfigure}[t]{0.17\textwidth}
       \includegraphics[width=\textwidth]{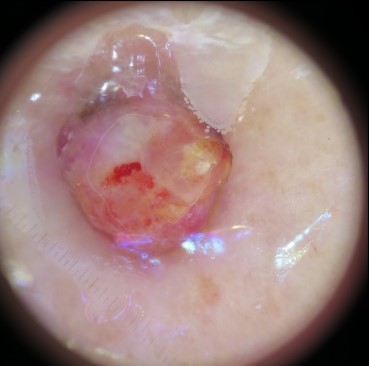}
       \caption{Original}
       \label{fig_no_ssl_a}
  \end{subfigure}
  ~
  \begin{subfigure}[t]{0.17\textwidth}
       \includegraphics[width=\textwidth]{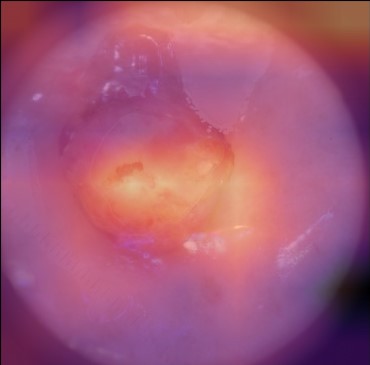}
       \caption{No pretraining}
       \label{fig_no_ssl_b}
  \end{subfigure} 
  ~
  \begin{subfigure}[t]{0.17\textwidth}
       \includegraphics[width=\textwidth]{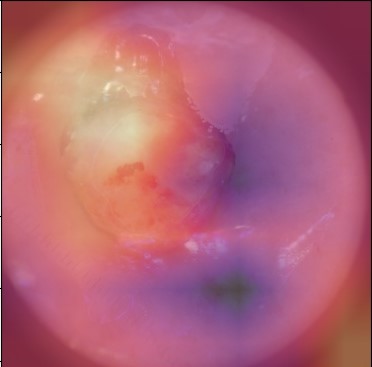}
       \caption{Supervised pretraining (ImageNet)}
       \label{fig_no_ssl_c}
  \end{subfigure}
  ~
  \begin{subfigure}[t]{0.17\textwidth}
       \includegraphics[width=\textwidth]{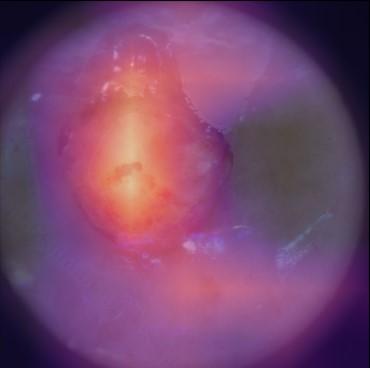}
       \caption{MoCo v2 (best single)}
       \label{fig_no_ssl_d}
  \end{subfigure}
  ~
  \begin{subfigure}[t]{0.17\textwidth}
       \includegraphics[width=\textwidth]{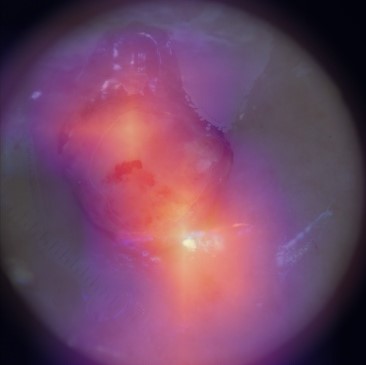}
       \caption{ODC $\,\to\,$ MoCo v2 (best combination)}
       \label{fig_no_ssl_e}
  \end{subfigure} 
  \caption{Class activation maps of classifiers with and without pretraining on pretext tasks. In the shown samples, classifiers that had no pretraining tend to focus on irrelevant parts of the images (black surrounding areas) and incorrectly classify skin lesions.}
  \label{fig_no_ssl}
\end{figure*}

\begin{figure*}[t]
  \centering
  \begin{subfigure}[t]{0.17\textwidth}
       \includegraphics[width=\textwidth]{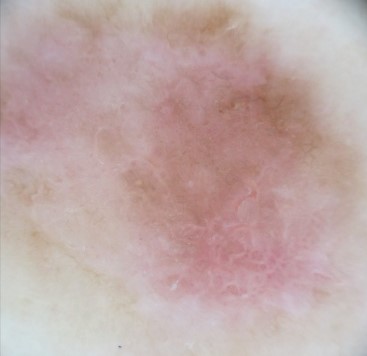}
  \end{subfigure}
  ~
  \begin{subfigure}[t]{0.17\textwidth}
       \includegraphics[width=\textwidth]{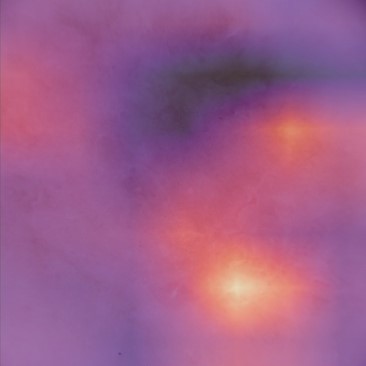}
  \end{subfigure}
  ~
  \begin{subfigure}[t]{0.17\textwidth}
       \includegraphics[width=\textwidth]{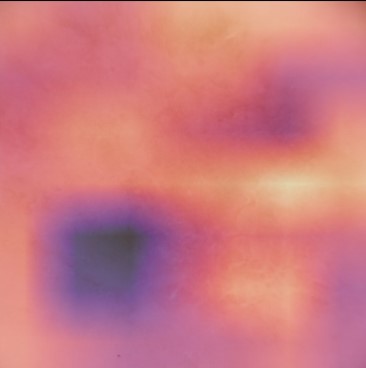}
  \end{subfigure} 
  ~
  \begin{subfigure}[t]{0.17\textwidth}
       \includegraphics[width=\textwidth]{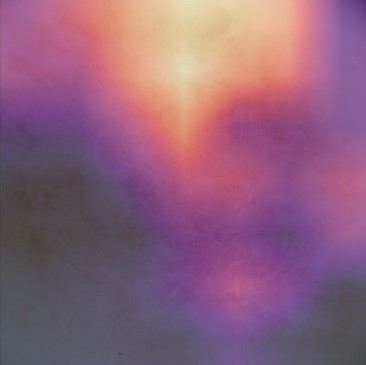}
  \end{subfigure} 
  ~
  \begin{subfigure}[t]{0.17\textwidth}
       \includegraphics[width=\textwidth]{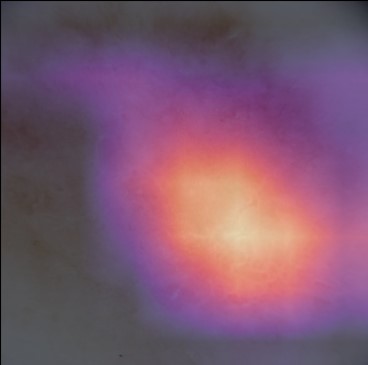}
  \end{subfigure}
  \hfill
  \begin{subfigure}[t]{0.17\textwidth}
       \includegraphics[width=\textwidth]{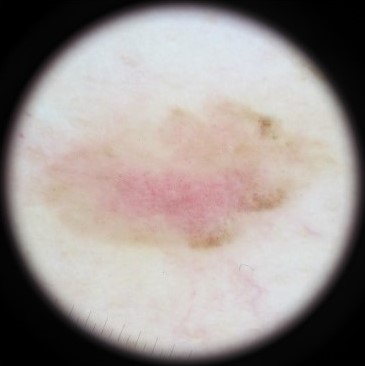}
       \caption{Original}
  \end{subfigure}
  ~
  \begin{subfigure}[t]{0.17\textwidth}
       \includegraphics[width=\textwidth]{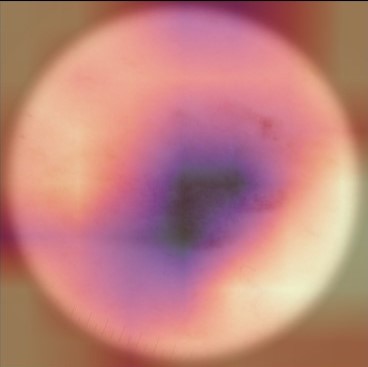}
       \caption{ODC}
  \end{subfigure} 
  ~
  \begin{subfigure}[t]{0.17\textwidth}
       \includegraphics[width=\textwidth]{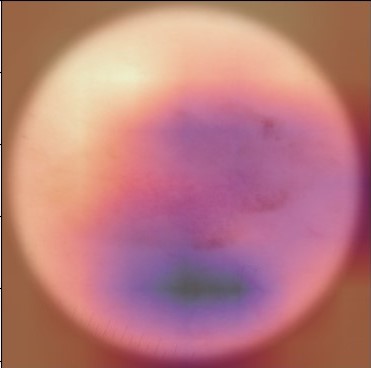}
       \caption{MoCo v2}
  \end{subfigure}
  ~
  \begin{subfigure}[t]{0.17\textwidth}
       \includegraphics[width=\textwidth]{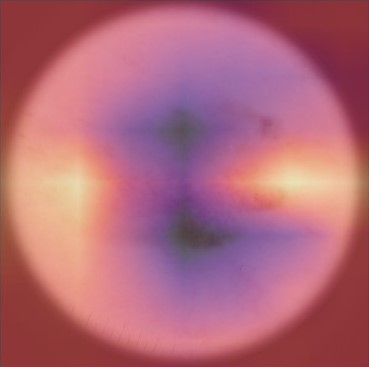}
       \caption{MoCo v2 $\,\to\,$ ODC (anti-curriculum)}
  \end{subfigure} 
  ~
  \begin{subfigure}[t]{0.17\textwidth}
       \includegraphics[width=\textwidth]{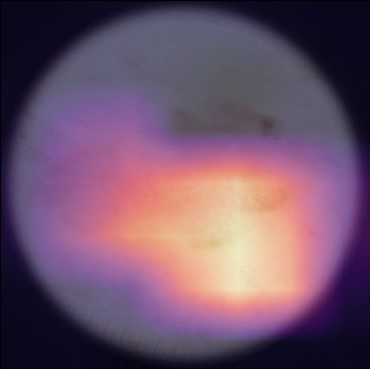}
       \caption{ODC $\,\to\,$ MoCo v2 (curriculum)}
  \end{subfigure}
  \caption{Class activation maps of classifiers pre-trained with ODC and/or MoCo v2. In the presented samples only the curriculum combination of pretext tasks correctly recognizes type of the lesion.
}
  \label{fig_single_tasks_vs_comb_odc_moco}
\end{figure*}

\begin{figure*}[t]
  \centering
  \begin{subfigure}[t]{0.17\textwidth}
       \includegraphics[width=\textwidth]{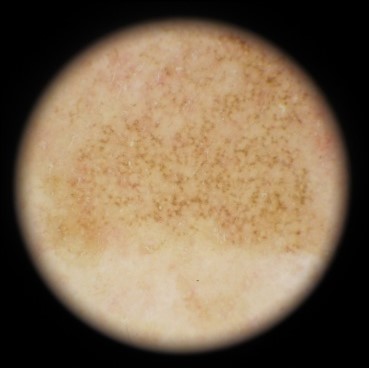}
  \end{subfigure}
  ~
  \begin{subfigure}[t]{0.17\textwidth}
       \includegraphics[width=\textwidth]{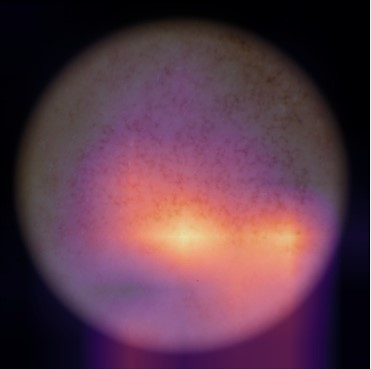}
  \end{subfigure}
  ~
  \begin{subfigure}[t]{0.17\textwidth}
       \includegraphics[width=\textwidth]{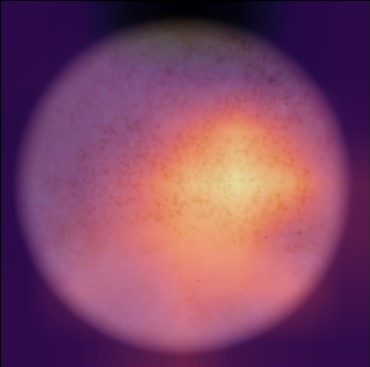}
  \end{subfigure} 
  ~
  \begin{subfigure}[t]{0.17\textwidth}
    \includegraphics[width=\textwidth]{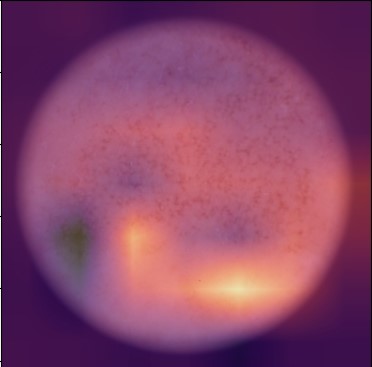}
  \end{subfigure} 
  ~
  \begin{subfigure}[t]{0.17\textwidth}
       \includegraphics[width=\textwidth]{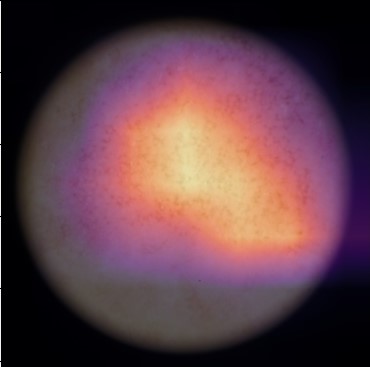}
  \end{subfigure}
  \hfill
  \begin{subfigure}[t]{0.17\textwidth}
       \includegraphics[width=\textwidth]{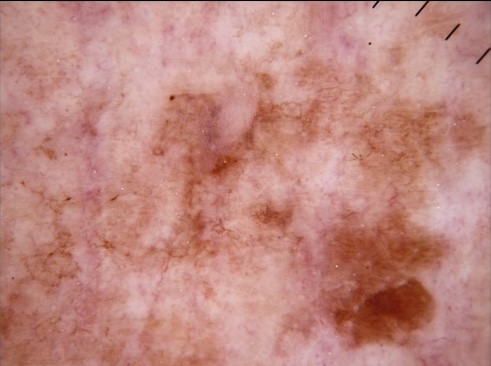}
       \caption{Original}
  \end{subfigure}
  ~
  \begin{subfigure}[t]{0.17\textwidth}
       \includegraphics[width=\textwidth]{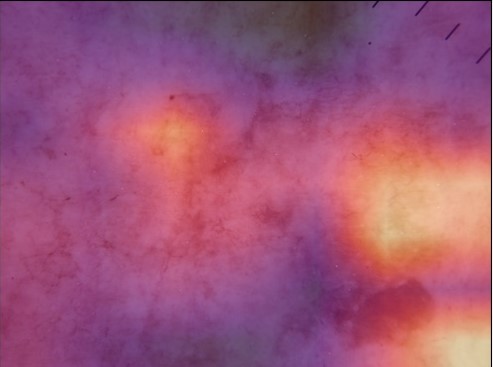}
       \caption{RL}
  \end{subfigure} 
  ~
  \begin{subfigure}[t]{0.17\textwidth}
       \includegraphics[width=\textwidth]{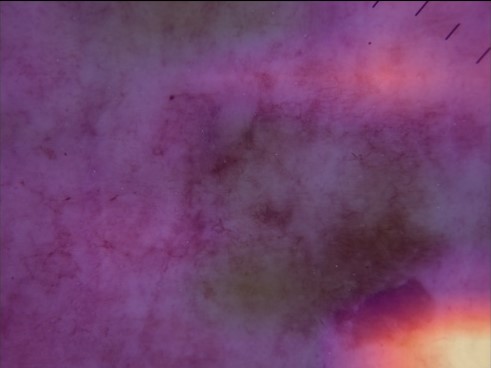}
       \caption{MoCo v2}
  \end{subfigure}
  ~
  \begin{subfigure}[t]{0.17\textwidth}
       \includegraphics[width=\textwidth]{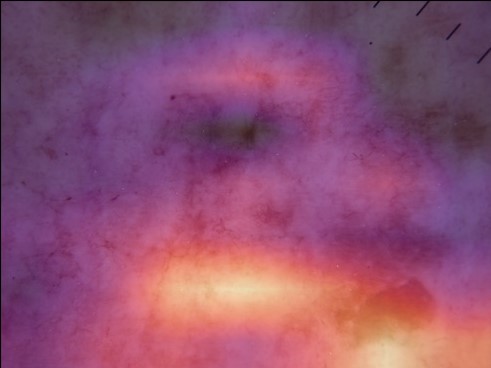}
       \caption{MoCo v2 $\,\to\,$~RL (anti-curriculum)}
  \end{subfigure} 
  ~
  \begin{subfigure}[t]{0.17\textwidth}
       \includegraphics[width=\textwidth]{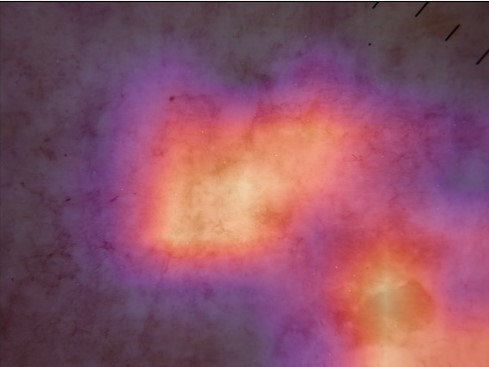}
       \caption{RL $\,\to\,$ MoCo v2 (curriculum)}
  \end{subfigure}
  \caption{Class activation maps of classifiers pre-trained with Relative Location and/or MoCo v2. In the presented samples only the curriculum combination of pretext tasks correctly recognizes type of the lesion. RL stands for Relative Location.}
  \label{fig_single_tasks_vs_comb_rl_moco}
\end{figure*}

Here, we use Class Activation Mapping (CAM)\cite{bib_cam} to demonstrate the advantages of a curriculum ordering of pretext tasks in terms of how accurately the model focuses on areas of the input image that are relevant to the target task. The use of CAM allows to visualizr the discriminative image areas that have a higher impact in the class prediction of a trained network.   






Figure \ref{fig_no_ssl} compares CAMs of models without pretraining, with those using supervised ImageNet or SSL pretraining, Figures \ref{fig_single_tasks_vs_comb_odc_moco} and  \ref{fig_single_tasks_vs_comb_rl_moco} present CAMs of models pretrained with two individual pretext tasks and their combinations arranged according to curriculum or anti-curriculum orderings: Figure \ref{fig_single_tasks_vs_comb_odc_moco} compares the effects of ODC and MoCo v2, and Figure \ref{fig_single_tasks_vs_comb_rl_moco}\textemdash Relative Location and MoCo v2 (we compare ODC and Relative Location against MoCo v2 as it results in the best individual performance).
Figure~\ref{fig_triplets} extends the analysis to the full curriculum and anti-curriculum orderings of pretext tasks.



Hereinafter, our visual analysis is based on the simple assumption that\textemdash for lesion assessment, the lesion and its borders are more important than other skin tissues (i.e. image regions that are not part of the lesion and its borders).

Figure \ref{fig_no_ssl} presents evidence that for the same input sample image, a non-pretrained model (Figure~\ref{fig_no_ssl_b}) does not focus attention on the skin lesion as much as the pretrained models do (Figures~\ref{fig_no_ssl_c}-\ref{fig_no_ssl_e})\textemdash CAM highlights indicate that a non-pretrained model focuses more attention on the parts of the images that do not contain lesions and gives more importance to the black areas surrounding the image that carry no useful information for image classification. Moreover, Figure \ref{fig_no_ssl_e} demonstrates that the best combination of pretext tasks leads to a model that focuses on the lesions better than the ImageNet-pretrained model: CAM highlights cover the lesion and are absent in other areas of the images (in the presented examples, only the SSL-pretrained models led to the correct classification of skin lesions). Figures~\ref{fig_single_tasks_vs_comb_odc_moco} and \ref{fig_single_tasks_vs_comb_rl_moco} further compare different SSL pretraining setups and show that curriculum ordering of pretext tasks significantly better focuses on the skin lesions than anti-curriculum ordering and individual tasks\textemdash in the CAM images the lesions are highlighted in brighter colors while the rest of the image is dark. In examples presented on Figures \ref{fig_single_tasks_vs_comb_odc_moco}~and~\ref{fig_single_tasks_vs_comb_rl_moco} only the curriculum ordering resulted in the correct classification of skin lesions.

\subsection{Computational complexity of multiple pretext task pretraining}

\setlength{\tabcolsep}{2pt}
\begin{table}
\centering
\caption{Hyperparameters used in the training of SSL and supervised models.}
\label{table_ssl_params}
\begin{tabular}{@{}ccccc@{}}
\toprule
 &
  \begin{tabular}[c]{@{}c@{}}Training\\ epochs\end{tabular} &
  \begin{tabular}[c]{@{}c@{}}Batch\\ size\end{tabular} &
  \begin{tabular}[c]{@{}c@{}}Learning\\ rate (LR)\end{tabular} &
  LR policy \\ \midrule
Relative location                                              & 70  & 64  & 0.2 & step {[}30; 50{]}                                          \\
MoCo v2                                                        & 200 & 32  & 0.03   & \begin{tabular}[c]{@{}c@{}}Cosine\\ Annealing\end{tabular} \\
ODC                                                            & 200 & 100 & 0.06   & step {[}170{]}                                             \\ \midrule
\begin{tabular}[c]{@{}c@{}}ImageNet\\ pretraining\end{tabular} & 90  & 256 & 0.1    & step {[}30; 60; 90{]}                                      \\
Classifier &
  90 &
  100 &
  0.1 &
  \begin{tabular}[c]{@{}c@{}}step {[}30; 60; 80{]},\\ warm-up: 10 epochs\\ warm-up ratio: 0.0001\end{tabular} \\
  \bottomrule
  \multicolumn{5}{p{251pt}}{The first three rows of the table present the hyperparameters used during pretraining by the SSL models. The last two rows show the hyperparameters used during the ImageNet-pretraining and the hyperparameters used during transferring to the target task.}
\end{tabular}
\end{table}

In this subsection we demonstrate that our proposed SSL-pretraining approach is computationally efficient and requires much less computational effort than typical supervised pretraining on ImageNet. We compare the computational complexity of different pretraining setups by comparing the number of iterations each setup requires to achieve the results indicated in Table~\ref{table_combinations_no_kd}, assuming that the impact on complexity associated with the differences in model architecture for different tasks (SSL or ImageNet-supervised heads) is negligible.

The number of iterations required to complete the training is expressed as:

\begin{equation}
    \mathcal{I} = N \cdot E / B,
\end{equation}
where $N$ is the number of training instances, $E$ is the number of training epochs and $B$ is the batch size. Therefore, for equal batch sizes, Relative Location requires $\mathcal{I}_{RL} = 7125$ iterations to train on 20264 instances (80\% of ISIC-2019), $\mathcal{I}_{ODC} = 15832$ iterations and $\mathcal{I}_{MoCo\_v2} = 15832$ iterations. On the other hand, supervised pretraining on ImageNet with the same batch size takes $\mathcal{I}_{ImageNet} = 421875$ iterations. Finally, transferring from an SSL model to the skin-lesion recognition task requires $\mathcal{I}_{transfer} = 7125$ iterations.

Pretraining a model on a sequence of curriculum ordered pretext tasks requires to: 
\begin{enumerate}
    \item Train all $k$ individual pretext tasks and perform single-source transfer learning from them to establish the curriculum.
    \item Train $k-1$ pretext tasks in the curriculum (the first one was already trained from scratch at the step one).
\end{enumerate}

Hence, the number of required iterations required:



\begin{equation} \label{eq_final}
    \mathcal{I}_{curriculum} = \mathcal{I}_{task_1} + 2\cdot\sum^{K}_{k=2} \mathcal{I}_{task_k} + K \cdot(\mathcal{I}_{transfer})
\end{equation} \label{eq_complexity}

In our case, $k=3$ (full curriculum). Thus, $\mathcal{I}_{curriculum} = \mathcal{I}_{ODC} + 2\cdot\mathcal{I}_{RL} + 2\cdot\mathcal{I}_{MoCo\_v2} + 3\cdot{\mathcal{I}_{transfer} = 83121}$, at most.

Therefore, the complexity of our full-curriculum approach $\mathcal{I}_{curriculum}$ is almost five times lower than time complexity of supervised ImageNet pretraining $I_{ImageNet} = 421875$. 

This complexity reduction is especially important for novel architectures that have not yet been pretrained on ImageNet, such as architectures that are tailor-designed, or automatically learned via Neural Architecture Search \cite{elsken2019neural}, to target a specific problem.





\section{Discussion}
\label{sec:Discussion}

It is common ground that transfer learing from a pretrained model usually leads to faster convergence and better performance than training from scratch on natural images~\cite{bib_higher_faster} or skin lesion images~\cite{bib_ssl_isic_2017}. This hypothesis is also confirmed for the models pretrained with SSL pretext tasks, as shown in Section~\ref{sec:Experiments}.


To our knowledge, we are the first studying the impact of the ordering of multiple sequential SSL tasks on overall performance gains: the full curriculum setup outperforms all three individual pretext tasks, while the full anti-curriculum setup sits between the two worst-performing single tasks. Interestingly, the highest performance results are obtained by a combination of just two pretext tasks (ODC and MoCo v2) in a curriculum ordering, instead of all three.


We hypothesize on the reasons behind the effects of different pretext task orderings in the following way; the fact that a pretext task leads to relatively low accuracy after transfer learning, may indicate that only basic pretext task specific features and representations have been learned. Differently, a pretext task resulting in higher downstream performances may be benefiting from the learning of generic features and representations. For this reason, if one of these last \textit{generically}-trained pretext tasks is followed by a \textit{specifically}-trained pretext task, the representations learned at the first stage may be harmed, as they will be countermanded to adapt for solving the specific pretext objective of the second task, loosing its generalization ability. Otherwise, if two pretext tasks result in generic features, their consecutive training is expected to either maintain or improve their individual performance. This reasoning allows to explain why curriculum (and also mixed curriculum) orderings perform better than anti-curriculum ones. 

A pair of pretext tasks ("Best curriculum" on Figure \ref{fig_accuracies}) outperforms a triplet of pretext tasks ("Full curriculum" on Figure \ref{fig_accuracies}), despite containing the same pretext tasks, ordered in the same way. Explicitly, the use of the previous specific-generic hypothesis comparing the best and full curriculum sequences would imply that the first (ODC) and the last (MoCo v2) tasks produce more effective and generic representations than the second task (Relative Location). Under this hypothesis, the presence of Relative Location might harm the overall performance, as MoCO v2 will not improve on the representations of ODC, but will try to \textit{recover} from the representations of Relative Location. However, Relative Location yields a higher downstream performance than ODC. Our hypothesis is that the training directions of ODC and MoCo v2 features are aligned, i.e, they both follow similar training paths, whereas intermediate Relative Location features in the full curriculum somehow \textit{shift} the feature space. 

\begin{figure}[t]
  \centering
       \includegraphics[width=0.5\textwidth]{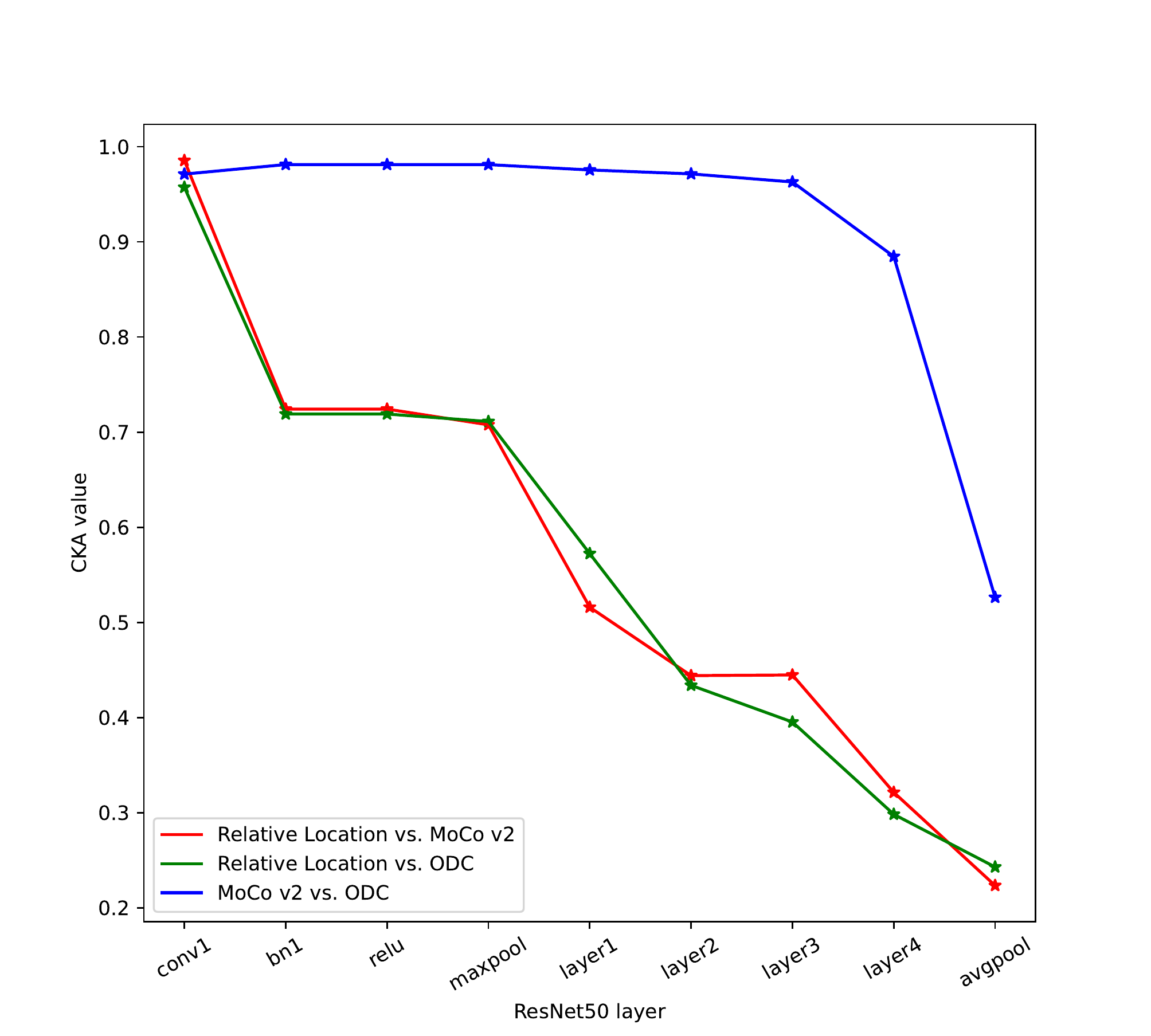}
       \caption{Similarities of feature maps learned by pretext tasks per layer. The similarity of feature maps is measured with Central Kernel Alignment (CKA): values of CKA closer to 1 indicate strong similarity of feature maps, while values closer to 0 indicate that they are dissimilar. }
  \label{fig_CKA}
\end{figure}

In order to test this hypothesis we conduct the following experiment: using the Central Kernel Alignment (CKA) \cite{bib_cka} to measure the similarity of representations at different network layers, we compare features learned by the pairs of individual pretext tasks as shown on Figure \ref{fig_CKA}. In accordance with our hypothesis: ODC and MoCo v2 produce very similar representations (the CKA values are close to one for all layers except the final average pooling layer), while the representations of Relative Location are very dissimilar from those of ODC and MoCo v2. This is also a strong evidence of the multi-path training idea, i.e. the fact that two models reaching similar downstream accuracy (as ODC and Relative Location) are relying on dissimilar representations.

This discussion also sheds light on the limitations of our approach: while the full curriculum ordering (where all pretext tasks are used) is better than any anti-curriculum and mixed curriculum orderings using all three tasks, it is not always the best option (which also hints that a simple increase of the number pretext tasks in the pretraining pipeline might not lead to better performance). However, we demonstrate that while it is not necessarily the best option, it shows competitive results and in our experiments outperforms the baselines. In this regard, as a direction for the future work, we suggest to incorporate the CKA analysis (alongside individual pretext tasks accuracy) to take into account the relationships between the learned representations of the individual tasks when establishing the curriculum ordering. 

\section{Conclusions}
\label{sec:Conclusions}

We explore the use of a sequential self-supervised pretraining pipeline to improve the accuracy of skin lesion classification from dermatoscopic images. We show that a model pretrained using a sequence of self-supervised pretext tasks on the target skin lesion dataset outperform models pretrained with individual pretext tasks, and a model pretrained on the ImageNet dataset in a supervised way. This approach also outperforms other top-performing solutions to the multi-class skin lesion recognition problem on the ISIC-2019 dataset, such as model ensembles. The qualitative analysis of the results shows that this strategy helps to better tune the focus of the model on the relevant parts of the image\textemdash skin lesion and its borders.

The experimental results also show that the order of the pretext tasks in the pretraining stage has a crucial effect on the accuracy on the downstream task, and that effective curriculum orderings in the pretext tasks correlate with increasing downstream accuracies obtained for individual pretext task pretraining. However, results also suggest that the optimal ordering are not solely related to the individual pretext task accuracy, but also the similarity in the representations obtained from different pretext tasks. Further work will explore the use of representation similarity metrics to establish optimal curriculum orderings in an automatic manner.  

Furthermore, our approach presents strong benefits for the computational complexity of the pretraining step, with respect to the ImageNet counterpart. It requires only a fraction of the ImageNet training time for relatively small datasets, such as ISIC-2019. This is  especially relevant for ad-hoc network   architectures for which a pretrained  ImageNet model may be unavailable.

\bibliography{bibli.bib}

\end{document}